\documentclass{INTERSPEECH2023}
\usepackage{algpseudocode}
\usepackage{algorithm}
\usepackage{amsmath}

\DeclareMathOperator*{\argminA}{arg\,min} % Jan Hlavacek
\DeclareMathOperator*{\argmaxA}{arg\,max} % Jan Hlavacek
\usepackage{comment}
\usepackage{bm}
\usepackage{xcolor}
\usepackage{multirow}
\usepackage{makecell}
\usepackage{textcomp}
\usepackage{graphicx}

\usepackage{makecell}
\usepackage{longtable}
\usepackage{subcaption}
\usepackage{lipsum}% http://ctan.org/pkg/lipsum
\usepackage{multicol}

\interspeechcameraready

\title{Mitigating the Exposure Bias in Sentence-Level Grapheme-to-Phoneme (G2P) Transduction}

\name{Eunseop Yoon$^1$\sthanks{\hspace{0.15cm}Equal contribution}, Hee Suk Yoon$^1$\footnotemark[1], Dhananjaya Gowda$^2$, SooHwan Eom$^1$, Daehyeok Kim$^1$, \\
John Harvill$^3$, Heting Gao$^3$, Mark Hasegawa-Johnson$^3$, Chanwoo Kim$^2$ ,and Chang D. Yoo$^1$\sthanks{\hspace{0.15cm}Corresponding author}}
\address{
  $^1$Korea Advanced Institute of Science and Technology (KAIST), Republic of Korea\\
  $^2$Samsung Electronics, Republic of Korea, 
  $^3$University of Illinois at Urbana-Champaign, USA}
\email{\{esyoon97, hskyoon, sean1105, kimshine, cd\_yoo\}@kaist.ac.kr, \\ 
        \{d.gowda, chanw.com\}@samsung.com, \{harvill2, hgao17, jhasegaw\}@illinois.edu}
        
\begin{document}

\maketitle
\begin{abstract}
% 1000 characters. ASCII characters only. No citations.
% Recently, ByT5 gave promising results on word-level G2P conversion by representing each input character with its corresponding UTF-8 encoding. Furthermore, a recent extension of T5, called ByT5, has been introduced as a token-free language model representing each character in a sequence with its corresponding UTF-8 encoding.  Previous research shows that ByT5 gives promising results on word-level G2P conversion, outperforming traditional token-based models.
%With the success of sequence-to-sequence capabilities of transformer architectures in various NLP tasks, Text-to-Text Transfer Transformer (T5) has been widely adapted for the Grapheme-to-Phoneme (G2P) conversion task. Recently, ByT5 gave promising results on word-level G2P conversion by representing each input character with its corresponding UTF-8 encoding. However, since ByT5 operates on the character level, it results in a higher number of decoding sequence steps required. This leads to a sub-optimal performance on the sentence-level G2P tasks, particularly as the sentence length increases, due to the exposure bias problem commonly observed in NLP models. In this paper, we show that the performance of sentence-level G2P can be improved by mitigating such exposure bias utilizing our proposed loss-based sampling method.
Text-to-Text Transfer Transformer (T5) has recently been considered for the Grapheme-to-Phoneme (G2P) transduction. As a follow-up, a tokenizer-free byte-level model based on T5 referred to as ByT5, recently gave promising results on word-level G2P conversion by representing each input character with its corresponding UTF-8 encoding. Although it is generally understood that sentence-level or paragraph-level G2P can improve usability in real-world applications as it is better suited to perform on heteronyms and linking sounds between words, we find that using ByT5 for these scenarios is nontrivial. Since ByT5 operates on the character level, it requires longer decoding steps, which deteriorates the performance due to the exposure bias commonly observed in auto-regressive generation models. This paper shows that the performance of sentence-level and paragraph-level G2P can be improved by mitigating such exposure bias using our proposed loss-based sampling method.

\end{abstract}

\section{Introduction}
%It is essential to develop a phonemic lexicon for TTS and ASR systems [1–4]. For this purpose, G2P techniques are used. For instance, modern TTS systems adopt G2P models as their frontend.
Grapheme-to-phoneme (G2P) transduction is essential for various applications that require phonetic lexicon, including automatic speech recognition (ASR) \cite{asr1} and text-to-speech synthesis (TTS) \cite{tts1}. There are two primary types of G2P tasks: word-level and sentence-level. In word-level G2P, the goal is to predict the pronunciation of a single word, whereas sentence-level G2P\footnote{For simplicity, sentence-level refers to both the sentence and paragraph-level (i.e., more than one sentence) G2P transduction.} involves predicting the pronunciations of all the words in a sentence. The latter is a more challenging task as it requires modeling context-dependent pronunciation variations of words (i.e., heteronyms) and linking sounds between words, making it more suitable for real-world applications. 

With the recent advancements in deep learning, transformer-based encoder-decoder language models, such as the Text-to-Text Transfer Transformer (T5) \cite{T5}, have emerged as a powerful tool for the G2P conversion \cite{tf,t5g2p,byt5g2p} where they learn to map input sequences (i.e., graphemes) to their corresponding output sequences (i.e., phonemes). Particularly, ByT5, a byte-level model based on T5, has been introduced as a token-free language model representing each character in a sequence with its corresponding UTF-8 encoding. Previous research shows that ByT5 gives promising results on word-level G2P conversion \cite{byt5g2p}, outperforming traditional token-based models. In this paper, we take a step towards extending ByT5 to the sentence-level G2P conversion. 

However, despite their remarkable performance, transformer-based models are affected by \textit{exposure bias}, which is a fundamental issue in auto-regressive generation models \cite{Ranzato2015SequenceLT}. This problem arises due to the discrepancy between the maximum likelihood training (also referred to as teacher forcing \cite{teacherforcing}) and the generation procedure during inference \cite{arora2022exposure}. As a result, errors can accumulate and propagate throughout the generation process, leading to a significant decrease in performance as the decoded sequence becomes longer.
%training and inference-time generation, where the model is trained on the ground-truth data distribution during training but generates the next token based on the sequences sampled from the model itself during inference-time. The distri- bution of these contexts seen during the generation phase might be very different from the ones encoun- tered during the training phase. As a result, errors can accumulate and propagate throughout the generation process, leading to a significant decrease in performance, particularly when the decoding sequence becomes longer. 

We show that due to the aforementioned exposure bias, the performance of ByT5 significantly deteriorates when applied to sentence-level G2P tasks due to the model's character-level operation, which leads to longer decoding sequences. This paper proposes a loss-dependent sampling method that builds upon a previously proposed two-pass decoding strategy \cite{scheduled} used to mitigate the exposure bias in Natural Language Processing (NLP). Our method involves identifying positions in the sequence with a high probability of wrong prediction by calculating the cross-entropy loss for each position. The positions with higher loss values are then sampled more frequently during training, in which their predictions are replaced into the ground truth phoneme sequence before inputting into the decoder. By sampling the positions in the sequence where errors are likely to occur during training, the proposed technique allows the model to learn from these errors and improve its ability to correct them. During this process, we use an adaptive sampling ratio determined by the Phoneme Error Rate (PER) of the previous epoch to determine the number of desired replacements adaptively.

With extensive experiments, we show that our loss-dependent sampling method improves the overall performance on our curated English G2P benchmark and quantitatively shows the reduction of exposure bias using the recently proposed metric \cite{arora2022exposure}. Moreover, we present some examples showing improved phoneme prediction due to the mitigation of exposure bias and improved heteronyms prediction over the word-level G2P ByT5.

\section{Related work}
\subsection{Grapheme-to-Phoneme (G2P) Conversion}
%Grapheme-to-phoneme conversion, also known as G2P, is the task of converting the sequence of the written unit symbols (which are called graphemes) into the corresponding sequence of the phonetic unit symbols (which are called phonemes). In other words, G2P aims to predict the pronunciation of a word or a sentence, given its spelling.
Early G2P models were based on a pronunciation dictionary, which looked up the corresponding pronunciation of the letter sequence. However, the size of the dictionary had to be very large and costly, and even then, they always had limited coverage with a finite-sized dictionary. Another early approach was the rule-based model, which determines the pronunciation of each letter or the subsequence based on a set of pre-defined phonetic rules \cite{rule1, rule2}. However, designing phonetic rules was often difficult, and still had difficulties capturing irregularities or complex rules, which are frequent in natural languages. To overcome the limitations of pre-defined phonetic rules, a data-driven stochastic approach that can mitigate complicated phonetic rules from the large dataset was widely used \cite{datadriven1, datadriven2, datadriven3}. These approaches usually involve weighted finite state transducers (WFST) \cite{wfst1, wfst2}.

Over time, the G2P conversion studies have shifted to deep learning-based methods \cite{tf,t5g2p,byt5g2p,cnn,lstm1,lstm2}. Recently, transformer-based language models have emerged as a powerful tool for the G2P conversion \cite{tf,t5g2p,byt5g2p} due to their effectiveness in modeling long-term dependencies in sequential data. Notably, the Text-to-Text Transfer Transformer (T5) \cite{T5}, which has gained attention due to its impressive performance on various NLP tasks, has been widely adapted for the G2P task \cite{t5g2p}.

\subsection{Token-Free Language Model}
Recently, token-free models that do not use word tokens have been emerging in NLP, and ByT5 \cite{byt5} has shown significantly better performance than traditional T5 models by using UTF-8 encoding as input. Because it does not use tokens for corresponding words or subwords, token-free models can process any language, even with unknown vocabulary. This makes it possible to handle a wide variety of languages with high performance, compared to mT5 \cite{mt5} using multilingual vocabulary embedding. Zhu et al. \cite{byt5g2p} has used this advantage for multilingual G2P conversion using ByT5.

\subsection{Exposure Bias}
Despite their significant achievements, transformer-based models suffer from exposure bias, a fundamental problem of auto-regressive natural language generation models \cite{Ranzato2015SequenceLT}. Exposure bias is explained to occur due to the discrepancy between training and test-time generation. During training, the model is trained based on the ground-truth data distribution. On the other hand, the model generates the next token during test time based on the prefix sequences sampled from the model itself. As a result, the error can propagate and accumulate throughout the generation process, leading to a critical degradation in performance, especially when the decoding sequence becomes longer \cite{schmidt-2019-generalization}. There have been several attempts to resolve exposure bias in auto-regressive sequence generation model. One way to tackle the problem is to replace or perturb the ground-truth sequence \cite{scheduled, scheduled2}. Some authors claim that the error accumulation can be solved with better decoding methods \cite{decoding1, decoding2}. Other attempts found the fault in maximum likelihood estimation (MLE) method and took non-MLE based approaches including reinforcement learning \cite{rl1, rl2} or generative adversarial networks \cite{gan}, while He et al. \cite{He2019ExposureBV} questions the impact of exposure bias in MLE training.

While it has been an active discussion on whether exposure bias really matters or not, Arora et al. \cite{arora2022exposure} have brought up a quantifiable definition of exposure bias and showed the accumulation of error due to the exposure bias actually exists during natural language generation.
%Previous work propose a two-pass decoding strategy where they mix the gold target sequence with the predicted sequence during training \cite{scheduled}.

\section{Loss-based Sampling}

\begin{figure}[t]
	\centering
    	\includegraphics[width=0.9\linewidth]{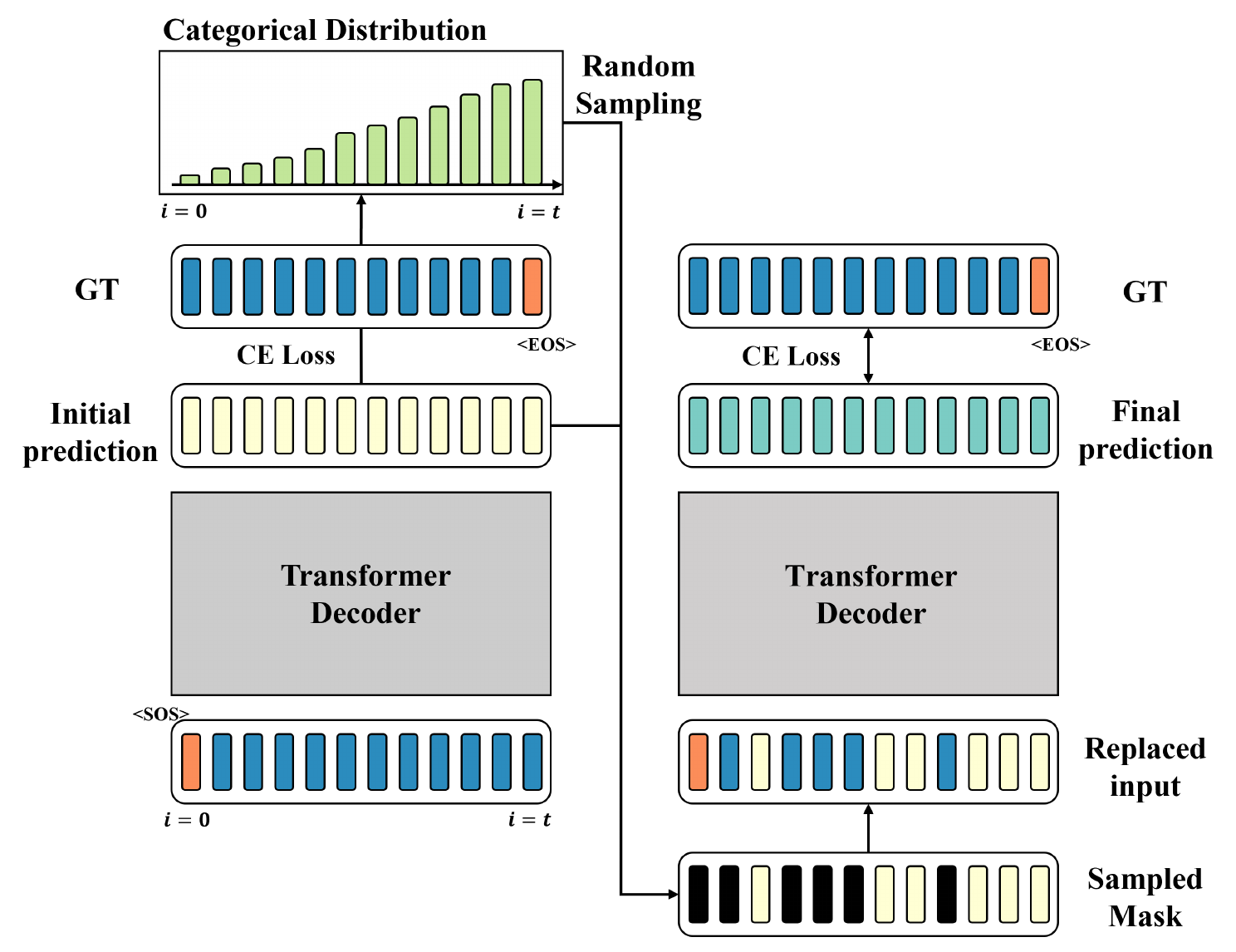}
	\caption{Our proposed loss-dependent sampling technique. The sampling method identifies positions in the sequence where errors are more likely to occur based on their cross-entropy loss values. By emphasizing these positions during training, the model is encouraged to pay more attention to them, resulting in improved accuracy of the predicted sequence during testing.}
	\label{fig:1}
\end{figure}
G2P is a sequence-to-sequence task where the input sequence (i.e., grapheme) $\textbf{X}=[x_0,x_1,x_2,...,x_n]$ is transformed to an output phoneme sequence $\textbf{Y}=[y_0,y_1,y_2,...,y_t]$ through a mapping model $p_{\theta}$, parameterized by $\theta$,  in which case we consider using the ByT5 \cite{byt5g2p}.

The training of $p_{\theta}$ is usually done via teacher forcing. The basic idea behind teacher forcing is to train the decoder to predict the next phoneme in a sequence given the previous phonemes (context) in that sequence. More formally, the model is trained by minimizing the negative log-likelihood for a given batch $B$,
\begin{equation}
\theta^{*} = \argminA_{\theta}\frac{-1}{|B|}\sum_{(\textbf{X},y_0^t) \in B} \sum_{i=0}^{t}{\log p_\theta(y_{i}|y_{0}^{i-1};\textbf{X})},
\end{equation}
where $y_{i}$ is the phoneme generated at step $i$, and $y_0^{i-1}$ is the context at time step $i$. During inference, however, the simplest strategy for generating a target sequence is to auto-regressively sample a sequence. That is, at each step $i$, pick the most probable phoneme $\hat{y}_i = \argmaxA p_{\theta}(\cdot|\hat{y}_0^{i-1};\textbf{X})$. The process carries on until either the sequence length limit is attained, or a unique token representing the end of the sequence (EOS) is produced.

Due to this discrepancy between the training objective and inference generation procedure, also referred to as exposure bias, auto-regressive generation models show degraded performance especially as the sequence of the decoding gets longer. 
%The basic idea behind teacher forcing is to train the decoder to predict the next word in a sequence given the previous words in that sequence,

\subsection{Loss-Based Sampling for Exposure Bias Mitigation}
Previous work proposed a sampling-based training which involves a two-pass decoding strategy where they randomly mix the gold target sequence with the predicted sequence to mitigate the exposure bias \cite{scheduled}. However, to give positions in the sequence where there is a high probability of wrong prediction during training, we propose a loss-based sampling method (Figure \ref{fig:1}).

The method involves several steps, starting with inputting the ground truth phoneme sequence and obtaining the predicted phoneme probability distribution for each time step $i \in [0,t]$, 
\begin{equation}
[p_{\theta}(\cdot|y_0^{-1};\textbf{X}), p_{\theta}(\cdot|y_0^{0};\textbf{X}),..., p_{\theta}(\cdot|y_0^{t-1};\textbf{X})].
%\argminA_{\theta}\frac{-1}{|B|}\sum_{(\textbf{X},y_0^t) \in B} \sum_{i=0}^{t}{\log p_\theta(y_{i}|y_{0}^{i-1};\textbf{X})},
\end{equation}
We then calculate the cross-entropy loss between each of the probability distributions with the one-hot encoding of the ground truth phoneme (note that we do not collect gradients during this step). 
\begin{equation}
%[D_{\text{KL}}(p_{0}^{-1}||o_0), D_{\text{KL}}(p_{0}^{0}||o_1),..., D_{\text{KL}}(p_{0}^{t-1}||o_t)],
[\text{H}(p_o,p_{\theta})_0, \text{H}(p_{o},p_{\theta})_1,..., \text{H}(p_{o},p_{\theta})_t],
%[D_{\text{KL}}(p_{\theta}(\cdot|y_0^{-1};\textbf{X}))|| \text{onehot}(y_0)), p_{\theta}(\cdot|y_0^{0};\textbf{X}), p_{\theta}(\cdot|y_0^{1};\textbf{X}), ..., p_{\theta}(\cdot|y_0^{t};\textbf{X})]
%\argminA_{\theta}\frac{-1}{|B|}\sum_{(\textbf{X},y_0^t) \in B} \sum_{i=0}^{t}{\log p_\theta(y_{i}|y_{0}^{i-1};\textbf{X})},
\end{equation}
where $\text{H}(p_{o},p_{\theta})_i = \text{CrossEntropy}(\text{onehot}(y_i),p_{\theta}(\cdot|y_0^{i-1};\textbf{X}) )$.

%Next, we normalize the KL divergences to obtain a categorical distribution. Using a specified 'sample ratio,' we then randomly sample elements from this distribution without replacement, creating a mask to determine whether to replace each Ground Truth input with its corresponding prediction. By implementing this method, we can obtain more accurate results in our study and mitigate the impact of exposure bias.
%Next, we obtain a categorical distribution based on the character-wise cross-entropy loss. 
Next, we normalize the cross-entropy loss to obtain a categorical distribution. From the distribution, we randomly sample a specific number of time steps without replacement, creating a mask to determine whether to replace each ground truth input with its corresponding prediction. During this sampling process, we use an \textit{adaptive sampling ratio} specified by the Phoneme Error Rate (PER) of the previous epoch to determine the number of desired replacements adaptively. The replaced sequence obtained from the loss-based sampling is once again used as the decoder input. Because the second prediction is generated based on its prediction randomly sampled, the resulting output can reflect the auto-regressive behavior. The cross-entropy loss calculated from the second prediction is our final loss and is used for backpropagation. 

The loss-based sampling technique we proposed allows the model to focus during training on positions in the sequence where errors are more likely to occur. By doing so, it can learn from these errors and improve its ability to correct them. This results in improved performance on longer inputs during testing, demonstrating the effectiveness of our approach.\label{sec:exposure_bias}

\subsection{Evaluation of Exposure Bias}

Arora et al. \cite{arora2022exposure} suggested using the time accumulation of the expected per-step prediction losses. During the evaluation, the decoder generates the predicted phoneme probability distribution $p_{\theta}(\hat{y}_i|\hat{y}_0^{i-1};\textbf{X})$ at each step $i$, based on the previous phoneme sequence $\hat{y}_0^{i-1}$ produced by the decoder. The predicted distribution is compared with the true distribution $p_{o}(\hat{y}_i|\hat{y}_0^{i-1};\textbf{X})$ to yield the per-step auto-regressive prediction loss:
\begin{align}
l^{AR}_{i}(\textbf{X}) & =D_{\text{KL}}(p_{o}(\hat{y}_i|\hat{y}_0^{i-1};\textbf{X})\,||\,p_{\theta}(\hat{y}_i|\hat{y}_0^{i-1};\textbf{X})) \\
& = \sum_{\hat{y}_i \in P} p_{o}(\hat{y}_i|\hat{y}_0^{i-1};\textbf{X}) \log{\frac{ p_{o}(\hat{y}_i|\hat{y}_0^{i-1};\textbf{X}) }{ p_{\theta}(\hat{y}_i|\hat{y}_0^{i-1};\textbf{X}) }},
\end{align}
where $P$ is the set of phonemes. The loss is averaged over the evaluation set $D_e$ to approximate the expected per-step auto-regressive loss:
\begin{align}
L^{AR}_{i} \approx \frac{1}{|D_e|} \sum_{(\textbf{X}, y_0^t) \in D_e} l^{AR}_{i}(\textbf{X}).
\end{align}
Similarly, the per-step teacher forcing losses are obtained by replacing $\hat{y}_0^{i-1}$ with the ground-truth phoneme sequence $y_0^{i-1}$, thereby assuming the absence of exposure bias,
\begin{gather}
L^{TF}_{i} \approx \frac{1}{|D_e|} \sum_{(\textbf{X}, y_0^t) \in D_e} l^{TF}_{i}(\textbf{X}, y_0^t), \\
l^{TF}_{i}(\textbf{X}, y_0^t)=D_{\text{KL}}(p_{o}(\hat{y}_i|y_0^{i-1};\textbf{X})\,||\,p_{\theta}(\hat{y}_i|y_0^{i-1};\textbf{X})).
\end{gather}
Then \cite{arora2022exposure} proposed the metric $\text{AccErr}_{\leq}(l)$, defined as the accumulation of the expected auto-regressive prediction loss relative to the expected teacher forcing loss until step $l$,
\begin{equation}
\text{AccErr}_{\leq}(l)=l \times \frac{\sum_{i=1}^{l} L^{AR}_{i}}{\sum_{i=1}^{l} L^{TF}_{i}}.
\end{equation}
The metric is known to have values between $l$ and $l^2$ under the proper assumption. The value of $l$ implies the absence of exposure bias for sequences of lengths up to $l$. On the other hand, a high deviation of $\text{AccErr}_{\leq}(l)$ from $l$ implies a high exposure bias for sequences of length $l$. In the worst-case scenario, the metric grows quadratically with the phoneme sequence length.

\section{Experimental Settings}

\subsection{Dataset and Model}
TIMIT \cite{timit} is a widely used corpus for speech processing tasks such as automatic speech recognition and text-to-speech. TIMIT contains 6300 spoken sentences, 10 by each of the 630 speakers from 8 major dialect regions of US English. TIMIT provides time-aligned orthographic and phonetic transcription, which can be used for labels in G2P conversion.
%dialect와 standard한 G2P mapping 차이 이야기?
%train/test split 이야기
In order to train and evaluate on the sentence-level G2P, we randomly concatenate up to 3 sentences during training. For the test set, we create two subsets, Testset$_{short}$ and Testset$_{long}$, which contain sentences concatenated up to 3 and 5, respectively. 
%Because the length of each transcription is not long enough to make length-dependent exposure bias visible, we concatenated up to 3 sentences randomly training and inference.
For the model selection, we used ByT5, a token-free model among variants of T5 \cite{byt5}. For all experiments, we finetune the pre-trained word-level G2P models\footnote{\url{https://github.com/lingjzhu/CharsiuG2P}}  \cite{byt5g2p} based on \texttt{ByT5-small}, which has about 300M parameters.

% To solve the exposure bias, mT5\footnote{\url{https://huggingface.co/google/mt5-small}}\cite{mt5}, a token-base T5 variant model, was used to reduce the length of input and output.

\subsection{Training Details}
We use AdamW optimizer \cite{AdamW} with $\beta_1=0.9$, $\beta_2=0.999$, $\epsilon=10^{-8}$,  and weight decay $\lambda=5 \times 10^{-3}$. The learning rate is set to $10^{-5}$, and the batch size is 32 for all experiments. We use gradient clipping with a 5.0 maximum gradient. The fixed sample ratio is chosen by grid search \{0,1.0.3, 0.6, 0.9\}, and the best ratio is independently selected for each experiment. The best model is selected by the lowest loss on the validation set. All experiments were done on NVIDIA Quadro RTX 8000.

\begin{table*}[t]
		\centering
	% \resizebox{\columnwidth}{!}{ % If your table exceeds the column or page width, use this command to reduce it slightly
	% \resizebox{\pagewidth}{!}{ % If your table exceeds the column or page width, use this command to reduce it slightly
	\begin{tabular}{ l c c c c c c c c c}
		\Xhline{2\arrayrulewidth}
            \multirow{3}{*}{Model  \textbackslash  Decoding Strategy} & \multicolumn{4}{c}{Testset$_{\textit{short}}$\text{ } (1-3 sentences)} &  & \multicolumn{4}{c}{Testset$_{\textit{long}}$\text{ } (4-5 sentences)}\\\cline{2-5}\cline{7-10}
		  & \multicolumn{2}{c}{Greedy search}& \multicolumn{2}{c}{Beam Search} & & \multicolumn{2}{c}{Greedy search}& \multicolumn{2}{c}{Beam Search}\\
         % & \begin{small}PER\end{small} & \begin{small}WER(\%)\end{small}  & \begin{small}PER(\%)\end{small} & \begin{small}WER(\%)\end{small}  & \begin{small}PER(\%)\end{small} & \begin{small}WER(\%)\end{small}  & \begin{small}PER(\%)\end{small} & \begin{small}WER(\%)\end{small}  \\
         & \begin{small}PER\end{small} & \begin{small}WER\end{small}  & \begin{small}PER\end{small} & \begin{small}WER\end{small} &   & \begin{small}PER\end{small} & \begin{small}WER\end{small}  & \begin{small}PER\end{small} & \begin{small}WER\end{small}  \\
		\Xhline{2\arrayrulewidth}
         \texttt{Teacher forcing(baseline)}&17.70 & 49.42  & 17.33& 48.54& & 24.10 & 55.66  & 23.65& 54.85\\
        \hline
        \textit{Training with sampling method} & & & &\\ \hline
            \textit{fixed sample ratio} \\
            $\cdot$ \begin{small}\texttt{Uniform sampling}\end{small}&16.07 & 48.69  & 15.95& 47.97& &22.25 & 54.40 & 21.98& 53.50\\
             $\cdot$ \begin{small}\texttt{Loss-based sampling(ours)}\end{small}&15.77 & 47.75  & 15.72&47.31&  & 22.65 & 55.81  & 21.73& 53.73\\
             \textit{adaptive sample ratio}\\
             $\cdot$ \begin{small}\texttt{Uniform sampling}\end{small}&16.18 & 47.94  & 15.71&46.90 &  &22.43 & 54.52  & 22.15& 53.67\\
             % $\cdot$ \texttt{Loss-based sampling (ours)}&16.94 & 48.62  & 15.66& 47.50&21.99 & 54.29  & 21.50& 53.81\\
             $\cdot$ \begin{small}\texttt{Loss-based sampling(ours)}\end{small}&\bf{15.64} & \bf{47.62}  & \bf{15.56} & \bf{46.50} & & \bf{21.99} & \bf{54.29}  & \bf{21.50}& \bf{53.41}\\

		\Xhline{3\arrayrulewidth} 

	\end{tabular}
	\caption{Comparison of PER (\%) and WER (\%) (lower is better) on the created test sets using TIMIT (Testset$_{short}$, Testset$_{long}$). \\For beam search, the number of beams is set to 3. The values reported are the average of three runs with different random seeds.}
	%}
	\label{tab:results}
\end{table*}

\begin{figure}[t]
	\centering
    	\includegraphics[width=0.86\linewidth]{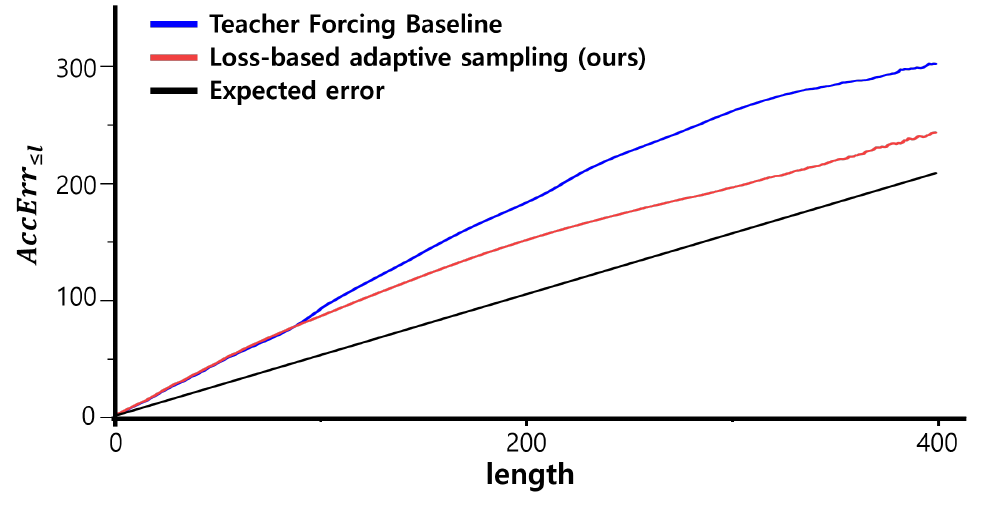}
	\caption{Plot of accumulated error up to length $l$ ($\text{AccErr}_{\leq}(l)=l$) w.r.t. $l$. The Expected Error curve represents the ideal case with no exposure bias.}
	\label{fig:2}
\end{figure}

\section{Results}

\begin{figure}[t]
	\centering
    	\includegraphics[width=\linewidth, height=6.5cm]{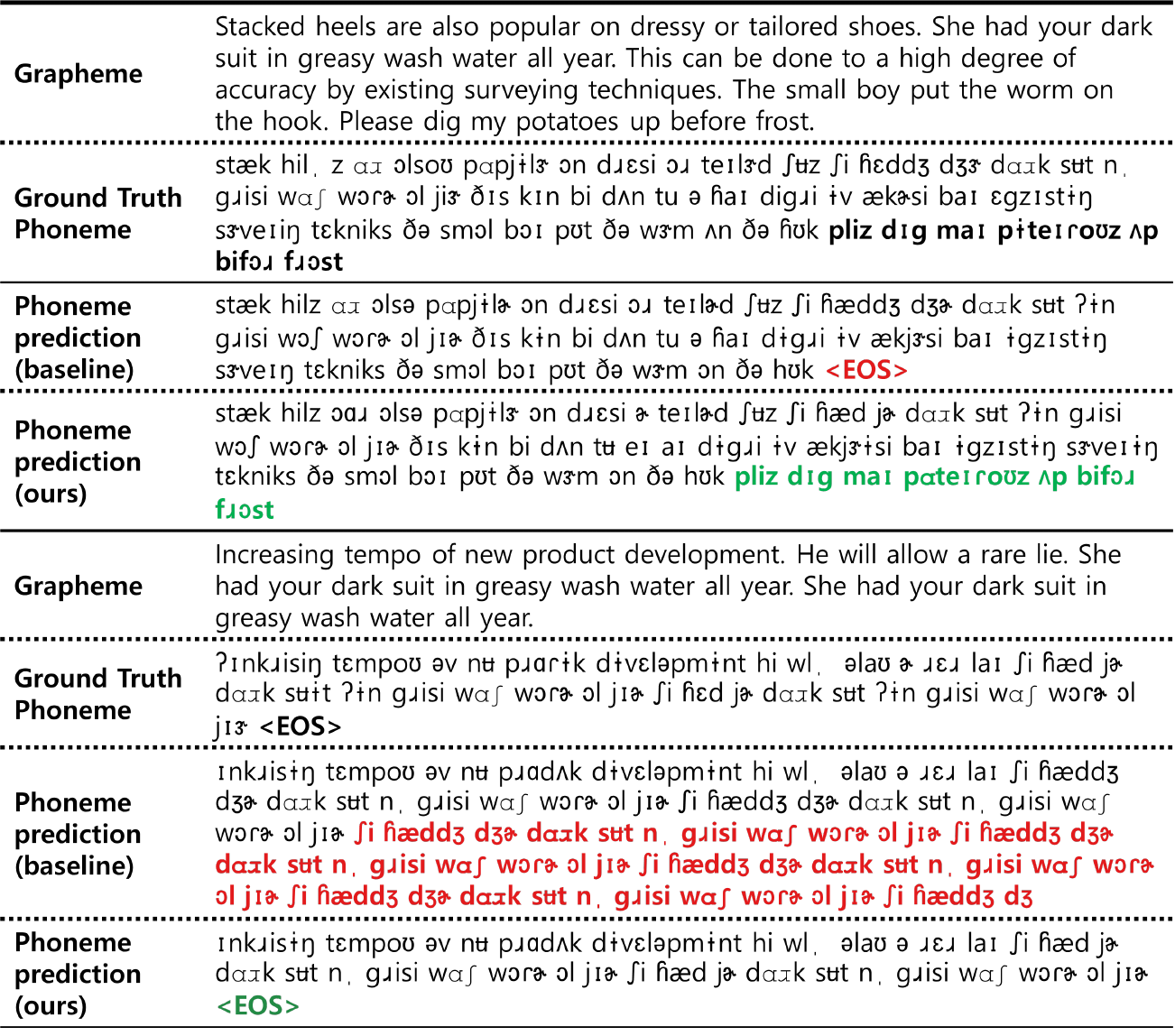}
	\caption{Examples of G2P conversion using greedy decoding for baseline and our method.} %In the first example, the baseline model fails to complete the prediction and terminates early, while our model predicts to the full sequence. In the second example, the baseline model continues generating phonemes beyond the true length of the sequence.}
	\label{fig:3}
\end{figure}

\subsection{Experiment Results}
Table \ref{tab:results} shows the Phoneme Error Rate (PER) and Word Error Rate (WER) of various sampling methods on different decoding strategies (greedy search and beam search) on the two groups of test set we created (i.e., Testset$_{short}$ and Testset$_{long}$). The naive two-pass decoding-based sampling method (uniform sampling) performs better than the teacher-forcing baseline. Our loss-based sampling method gives better results than the uniform sampling for both the fixed and adaptive sample ratio settings. Our loss-based sampling with adaptive sample ratio on Testset$_{long}$ obtained 21.99\% with greedy search and 21.50\% with beam search, outperforming teacher forcing baseline by 2.11\%p and 2.15\%p, respectively.

In Figure \ref{fig:2}, we analyze the exposure bias on teacher forcing baseline and our loss-based adaptive sampling method by plotting the accumulated error with the expected error base on Section \ref{sec:exposure_bias}. The difference with the expected error is significantly reduced compared to the baseline due to our loss-based sampling method effectively mitigating the exposure bias.
\subsection{Comparison of G2P Quality}

Figure \ref{fig:3} shows a comparison of the phoneme predictions for a given Grapheme between a baseline and our model. In the first example, the baseline model fails to complete the prediction and terminates early, while our proposed model accurately predicts the full sequence. This demonstrates the superiority of our model in handling long input sequences and producing more accurate predictions. In the second example, the baseline model continues generating phonemes beyond the true length of the sequence while our model produces the correct number of phonemes. This further confirms the effectiveness of our proposed model by avoiding unnecessary phoneme generation.

\begin{figure}[t]
	\centering
    	\includegraphics[width=\linewidth, height=2.5cm]{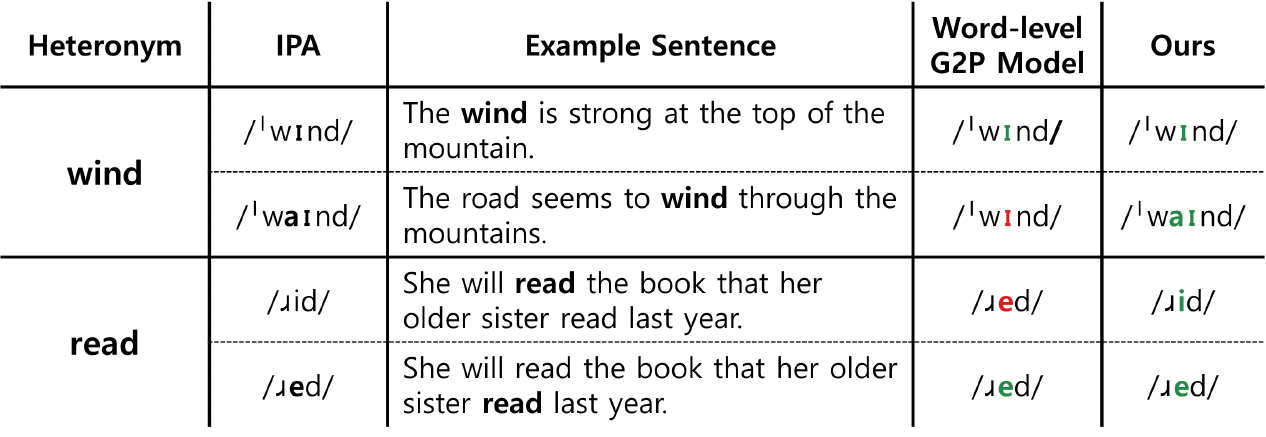}
	\caption{Examples of heteronyms predictions using word-level and sentence-level G2P system (Ours).}
	\label{fig:4}
\end{figure}

\subsection{Heteronyms: Context Dependency}

Heteronyms are words that are spelled the same but have different pronunciations and meanings depending on the context. Word-level G2P systems cannot understand the context in which the word is used, limiting their capability of disambiguating heteronyms. On the other hand, sentence-level G2P systems can analyze the whole input sequence to determine the correct pronunciation for heteronyms. Figure \ref{fig:4} illustrates some examples of heteronym predictions in which it demonstrates that the previously proposed word-level ByT5 \cite{byt5g2p} is unable to disambiguate the phoneme in contrast to our sentence-level ByT5.

\section{Conclusions}
We find that ByT5 G2P performance for longer decoding sequences in sentence or paragraph-level G2P tasks is sub-optimal due to exposure bias. However, the proposed loss-based sampling method has been found to mitigate this problem and improve the performance of sentence and paragraph-level G2P using ByT5. These findings highlight the potential for further improvements in G2P conversion using transformer architectures. 

\section{Limitation}
Our research on sentence and paragraph-level G2P conversion was limited by the need for high-quality phoneme-labeled data incorporating contextual information and linking sounds. The TIMIT dataset has human-labeled phonemes for text utterances but may reflect pronunciation specific to particular dialects. Despite these limitations, our study proposes a novel approach to sentence-level G2P that can address some of these challenges and provides valuable insights for future research.
%In the case of word-level g2p, gold standard data can be collected through a dictionary, but in the case of sentence-level g2p, if data is collected in the same way, it cannot contain enough information about the heteronym according to the context and linking sound in the sentence. Among the open data sets, an experiment was conducted only on TIMIT,

\section{Acknowledgements}
This work was supported by Institute for Information \& communications Technology Planning \& Evaluation (IITP) grant funded by the Korea government(MSIT) (No. 2021-0-01381) and Institute of Information \& communications Technology Planning \& Evaluation (IITP) grant funded by the Korea government(MSIT) (No.2022-0-00184, Development and Study of AI Technologies to Inexpensively Conform to Evolving Policy on Ethics).

\begin{comment}
\section{Acknowledgements}

\ifinterspeechfinal
     The INTERSPEECH 2023 organisers
\else
     The authors
\fi
would like to thank ISCA and the organising committees of past INTERSPEECH conferences for their help and for kindly providing the previous version of this template.

As a final reminder, the 5th page is reserved exclusively for references. No other content must appear on the 5th page. Appendices, if any, must be within the first 4 pages. The references may start on an earlier page, if there is space.
\end{comment}
% \newpage

\bibliographystyle{IEEEtran}
\bibliography{main}

% Generated by IEEEtran.bst, version: 1.13 (2008/09/30)
\begin{thebibliography}{10}
\providecommand{\url}[1]{#1}
\csname url@samestyle\endcsname
\providecommand{\newblock}{\relax}
\providecommand{\bibinfo}[2]{#2}
\providecommand{\BIBentrySTDinterwordspacing}{\spaceskip=0pt\relax}
\providecommand{\BIBentryALTinterwordstretchfactor}{4}
\providecommand{\BIBentryALTinterwordspacing}{\spaceskip=\fontdimen2\font plus
\BIBentryALTinterwordstretchfactor\fontdimen3\font minus
  \fontdimen4\font\relax}
\providecommand{\BIBforeignlanguage}[2]{{%
\expandafter\ifx\csname l@#1\endcsname\relax
\typeout{** WARNING: IEEEtran.bst: No hyphenation pattern has been}%
\typeout{** loaded for the language `#1'. Using the pattern for}%
\typeout{** the default language instead.}%
\else
\language=\csname l@#1\endcsname
\fi
#2}}
\providecommand{\BIBdecl}{\relax}
\BIBdecl

\bibitem{asr1}
M.~A. Hasegawa-Johnson, L.~Rolston, C.~Goudeseune, G.-A. Levow, and
  K.~Kirchhoff, ``Grapheme-to-phoneme transduction for cross-language asr,'' in
  \emph{International Conference on Statistical Language and Speech
  Processing}, 2020.

\bibitem{tts1}
Z.~Hong, J.~Wang, X.~Qu, J.~Liu, C.~Zhao, and J.~Xiao, ``{Federated Learning
  with Dynamic Transformer for Text to Speech},'' in \emph{Proc. Interspeech
  2021}, 2021, pp. 3590--3594.

\bibitem{T5}
\BIBentryALTinterwordspacing
C.~Raffel, N.~Shazeer, A.~Roberts, K.~Lee, S.~Narang, M.~Matena, Y.~Zhou,
  W.~Li, and P.~J. Liu, ``Exploring the limits of transfer learning with a
  unified text-to-text transformer,'' \emph{CoRR}, vol. abs/1910.10683, 2019.
  [Online]. Available: \url{http://arxiv.org/abs/1910.10683}
\BIBentrySTDinterwordspacing

\bibitem{tf}
S.~Yolchuyeva, G.~N{\'e}meth, and B.~Gyires-T{\'o}th, ``Transformer based
  grapheme-to-phoneme conversion,'' in \emph{Interspeech}, 2019.

\bibitem{t5g2p}
M.~Rez{\'a}ckov{\'a}, J.~Svec, and D.~Tihelka, ``T5g2p: Using text-to-text
  transfer transformer for grapheme-to-phoneme conversion,'' in
  \emph{Interspeech}, 2021.

\bibitem{byt5g2p}
J.~Zhu, C.~Zhang, and D.~Jurgens, ``Byt5 model for massively multilingual
  grapheme-to-phoneme conversion,'' in \emph{Interspeech}, 2022.

\bibitem{Ranzato2015SequenceLT}
M.~Ranzato, S.~Chopra, M.~Auli, and W.~Zaremba, ``Sequence level training with
  recurrent neural networks,'' \emph{CoRR}, vol. abs/1511.06732, 2015.

\bibitem{teacherforcing}
R.~J. Williams and D.~Zipser, ``A learning algorithm for continually running
  fully recurrent neural networks,'' \emph{Neural Computation}, vol.~1, no.~2,
  pp. 270--280, 1989.

\bibitem{arora2022exposure}
K.~Arora, L.~E. Asri, H.~Bahuleyan, and J.~C.~K. Cheung, ``Why exposure bias
  matters: An imitation learning perspective of error accumulation in language
  generation,'' \emph{arXiv preprint arXiv:2204.01171}, 2022.

\bibitem{scheduled}
\BIBentryALTinterwordspacing
T.~Mihaylova and A.~F.~T. Martins, ``Scheduled sampling for transformers,'' in
  \emph{Proceedings of the 57th Annual Meeting of the Association for
  Computational Linguistics: Student Research Workshop}.\hskip 1em plus 0.5em
  minus 0.4em\relax Florence, Italy: Association for Computational Linguistics,
  Jul. 2019, pp. 351--356. [Online]. Available:
  \url{https://aclanthology.org/P19-2049}
\BIBentrySTDinterwordspacing

\bibitem{rule1}
H.~S. Elovitz, R.~W. Johnson, A.~McHugh, and J.~E. Shore, ``Automatic
  translation of english text to phonetics by means of letter-to-sound rules,''
  NAVAL RESEARCH LAB WASHINGTON DC, Tech. Rep., 1976.

\bibitem{rule2}
R.~M. Kaplan and M.~Kay, ``Regular models of phonological rule systems,''
  \emph{Comput. Linguist.}, vol.~20, no.~3, p. 331–378, sep 1994.

\bibitem{datadriven1}
R.~I. Damper, Y.~Marchand, M.~J. Adamson, and K.~Gustafson, ``Comparative
  evaluation of letter-to-sound conversion techniques for english
  text-to-speech synthesis,'' in \emph{Speech Synthesis Workshop}, 1998.

\bibitem{datadriven2}
M.~Bisani and H.~Ney, ``Investigations on joint-multigram models for
  grapheme-to-phoneme conversion,'' \emph{7th International Conference on
  Spoken Language Processing (ICSLP 2002)}, 2002.

\bibitem{datadriven3}
M.~Razavi, R.~Rasipuram, and M.~Magimai.-Doss, ``Acoustic data-driven
  grapheme-to-phoneme conversion in the probabilistic lexical modeling
  framework,'' \emph{Speech Commun.}, vol.~80, pp. 1--21, 2016.

\bibitem{wfst1}
D.~Caseiro, I.~Trancoso, L.~O. Inesc-Id, Ist, R.~A. Redol, and C.~Viana,
  ``Grapheme-to-phone using finite-state transducers,'' \emph{Proceedings of
  2002 IEEE Workshop on Speech Synthesis, 2002.}, pp. 215--218, 2002.

\bibitem{wfst2}
\BIBentryALTinterwordspacing
J.~R. Novak, N.~Minematsu, and K.~Hirose, ``{WFST}-based grapheme-to-phoneme
  conversion: Open source tools for alignment, model-building and decoding,''
  in \emph{Proceedings of the 10th International Workshop on Finite State
  Methods and Natural Language Processing}.\hskip 1em plus 0.5em minus
  0.4em\relax Donostia{--}San Sebasti{\'a}n: Association for Computational
  Linguistics, Jul. 2012, pp. 45--49. [Online]. Available:
  \url{https://aclanthology.org/W12-6208}
\BIBentrySTDinterwordspacing

\bibitem{cnn}
S.~Yolchuyeva, G.~Németh, and B.~Gyires-Tóth, ``Grapheme-to-phoneme
  conversion with convolutional neural networks,'' \emph{Applied Sciences},
  vol.~9, p. 1143, 03 2019.

\bibitem{lstm1}
K.~Rao, F.~Peng, H.~Sak, and F.~Beaufays, ``Grapheme-to-phoneme conversion
  using long short-term memory recurrent neural networks,'' in \emph{2015 IEEE
  International Conference on Acoustics, Speech and Signal Processing
  (ICASSP)}, 2015, pp. 4225--4229.

\bibitem{lstm2}
\BIBentryALTinterwordspacing
K.~Yao and G.~Zweig, ``Sequence-to-sequence neural net models for
  grapheme-to-phoneme conversion,'' \emph{CoRR}, vol. abs/1506.00196, 2015.
  [Online]. Available: \url{http://arxiv.org/abs/1506.00196}
\BIBentrySTDinterwordspacing

\bibitem{byt5}
\BIBentryALTinterwordspacing
L.~Xue, A.~Barua, N.~Constant, R.~Al-Rfou, S.~Narang, M.~Kale, A.~Roberts, and
  C.~Raffel, ``{B}y{T}5: Towards a token-free future with pre-trained
  byte-to-byte models,'' \emph{Transactions of the Association for
  Computational Linguistics}, vol.~10, pp. 291--306, 2022. [Online]. Available:
  \url{https://aclanthology.org/2022.tacl-1.17}
\BIBentrySTDinterwordspacing

\bibitem{mt5}
\BIBentryALTinterwordspacing
L.~Xue, N.~Constant, A.~Roberts, M.~Kale, R.~Al-Rfou, A.~Siddhant, A.~Barua,
  and C.~Raffel, ``m{T}5: A massively multilingual pre-trained text-to-text
  transformer,'' in \emph{Proceedings of the 2021 Conference of the North
  American Chapter of the Association for Computational Linguistics: Human
  Language Technologies}.\hskip 1em plus 0.5em minus 0.4em\relax Online:
  Association for Computational Linguistics, Jun. 2021, pp. 483--498. [Online].
  Available: \url{https://aclanthology.org/2021.naacl-main.41}
\BIBentrySTDinterwordspacing

\bibitem{schmidt-2019-generalization}
\BIBentryALTinterwordspacing
F.~Schmidt, ``Generalization in generation: A closer look at exposure bias,''
  in \emph{Proceedings of the 3rd Workshop on Neural Generation and
  Translation}.\hskip 1em plus 0.5em minus 0.4em\relax Hong Kong: Association
  for Computational Linguistics, Nov. 2019, pp. 157--167. [Online]. Available:
  \url{https://aclanthology.org/D19-5616}
\BIBentrySTDinterwordspacing

\bibitem{scheduled2}
S.~Bengio, O.~Vinyals, N.~Jaitly, and N.~Shazeer, ``Scheduled sampling for
  sequence prediction with recurrent neural networks,'' in \emph{Proceedings of
  the 28th International Conference on Neural Information Processing Systems -
  Volume 1}, ser. NIPS'15.\hskip 1em plus 0.5em minus 0.4em\relax Cambridge,
  MA, USA: MIT Press, 2015, p. 1171–1179.

\bibitem{decoding1}
\BIBentryALTinterwordspacing
S.~Welleck, I.~Kulikov, J.~Kim, R.~Y. Pang, and K.~Cho, ``Consistency of a
  recurrent language model with respect to incomplete decoding,'' in
  \emph{Proceedings of the 2020 Conference on Empirical Methods in Natural
  Language Processing (EMNLP)}.\hskip 1em plus 0.5em minus 0.4em\relax Online:
  Association for Computational Linguistics, Nov. 2020, pp. 5553--5568.
  [Online]. Available: \url{https://aclanthology.org/2020.emnlp-main.448}
\BIBentrySTDinterwordspacing

\bibitem{decoding2}
R.~Collobert, A.~Y. Hannun, and G.~Synnaeve, ``A fully differentiable beam
  search decoder,'' \emph{ArXiv}, vol. abs/1902.06022, 2019.

\bibitem{rl1}
Y.~Chen, L.~Wu, and M.~J. Zaki, ``Reinforcement learning based
  graph-to-sequence model for natural question generation,'' \emph{ArXiv}, vol.
  abs/1908.04942, 2019.

\bibitem{rl2}
D.~Bahdanau, P.~Brakel, K.~Xu, A.~Goyal, R.~Lowe, J.~Pineau, A.~C. Courville,
  and Y.~Bengio, ``An actor-critic algorithm for sequence prediction,''
  \emph{ArXiv}, vol. abs/1607.07086, 2016.

\bibitem{gan}
L.~Yu, W.~Zhang, J.~Wang, and Y.~Yu, ``Seqgan: Sequence generative adversarial
  nets with policy gradient,'' in \emph{Proceedings of the Thirty-First AAAI
  Conference on Artificial Intelligence}, ser. AAAI'17.\hskip 1em plus 0.5em
  minus 0.4em\relax AAAI Press, 2017, p. 2852–2858.

\bibitem{He2019ExposureBV}
T.~He, J.~Zhang, Z.~Zhou, and J.~R. Glass, ``Exposure bias versus
  self-recovery: Are distortions really incremental for autoregressive text
  generation?'' in \emph{Conference on Empirical Methods in Natural Language
  Processing}, 2019.

\bibitem{timit}
C.~Lopes and F.~Perdig{\~a}o, ``Timit acoustic-phonetic continuous speech
  corpus,'' 2012.

\bibitem{AdamW}
\BIBentryALTinterwordspacing
I.~Loshchilov and F.~Hutter, ``Decoupled weight decay regularization,'' in
  \emph{International Conference on Learning Representations}, 2019. [Online].
  Available: \url{https://openreview.net/forum?id=Bkg6RiCqY7}
\BIBentrySTDinterwordspacing

\end{thebibliography}

\end{document}